\documentclass{article}

     \PassOptionsToPackage{numbers, compress}{natbib}

     \usepackage[preprint]{neurips_2022}

\usepackage[utf8]{inputenc} 
\usepackage[T1]{fontenc}    
\usepackage{hyperref}       
\usepackage{url}            
\usepackage{booktabs}       
\usepackage{amsfonts}       
\usepackage{nicefrac}       
\usepackage{microtype}      
\usepackage{xcolor}         
\usepackage{siunitx}
\usepackage{amsmath,amssymb}
\usepackage{graphicx}
\usepackage{subcaption}
\usepackage{caption} 
\usepackage{adjustbox}

\DeclareMathOperator{\EX}{\mathbb{E}}

\title{Amortized Bayesian Inference of GISAXS Data with Normalizing Flows}

\author{%
  Maksim Zhdanov \\
  Helmholtz-Zentrum Dresden-Rossendorf \\
  \texttt{maxxxzdn@gmail.com} \\
   \And
   Lisa Randolph \\
   European XFEL \\
   \AND
   Thomas Kluge \\
   Helmholtz-Zentrum Dresden-Rossendorf \\
   \And
   Motoaki Nakatsutsumi \\
   European XFEL \\
   \And
   Christian Gutt \\
   University of Siegen \\
   \AND
   Marina Ganeva \\
   Forschungszentrum Jülich \\
   \And
   Nico Hoffmann \\
   Helmholtz-Zentrum Dresden-Rossendorf
}

\begin{document}

\maketitle

\begin{abstract}
Grazing-Incidence Small-Angle X-ray Scattering (GISAXS) is a modern imaging technique used in material research to study nanoscale materials. Reconstruction of the parameters of an imaged object imposes an ill-posed inverse problem that is further complicated when only an in-plane GISAXS signal is available. Traditionally used inference algorithms such as Approximate Bayesian Computation (ABC) rely on computationally expensive scattering simulation software, rendering analysis highly time-consuming. We propose a simulation-based framework that combines variational auto-encoders and normalizing flows to estimate the posterior distribution of object parameters given its GISAXS data. We apply the inference pipeline to experimental data and demonstrate that our method reduces the inference cost by orders of magnitude while producing consistent results with ABC.
\end{abstract}

\section{Introduction}

\paragraph{X-ray scattering}
Structural and morphological properties of surfaces and multi-layer thin films can be investigated by grazing-incidence small angle X-ray scattering (GISAXS). An incident X-ray beam hits the sample under grazing incidence, after which its scattering pattern is recorded by a 2D area detector. The analysis of GISAXS patterns allows one to observe in-situ transient surface and subsurface density profiles with nanometer depth resolution. It is important for studying the growth of nanomaterials, as demonstrated by Metwalli et al. \cite{Metwalli2013}, and for investigating the (sub-)picosecond surface dynamics of laser-irradiated solids, as shown by Randolph et al. \cite{Lisa2022}. The reconstruction of the structural properties is complicated by multiple scattering contributions and the missing phase information. As the number of layers grows, the scattering patterns get highly complex, rendering the inference computationally infeasible without prior knowledge about imaged objects.

\paragraph{Related work}
Traditionally, a simulated GISAXS pattern of a sample model is generated and compared with the experimentally measured pattern. Iterative fitting processes adjust the model parameters by minimizing $\chi^2$ until they match the experimental GISAXS pattern \cite{Lisa2022, Ukleev2019}. 
The need to execute a forward model iteratively makes a comprehensive analysis of GISAXS images highly time-consuming, which in turn limits the throughput of large-scale neutron and X-ray facilities. The need for accelerated inference motivated the spread of machine learning-based algorithms. Cherukara et al. \cite{Cherukara2018} demonstrated the real-time inversion of Wide-Angle X-ray scattering using generative neural networks yielding 500 times faster inference compared to standard iterative algorithms. Ikemoto et al. \cite{Ikemoto2020} and Liu et al. \cite{Liu2019} use convolutional neural networks for the one-step classification of experimental images. At the same time, Van Herck et al. \cite{VanHerck2021} infer the rotation distribution of nanoparticle arrangements. Similar to our use case, Mironov et al. \cite{Mironov2021} use convolutional neural networks with uncertainty quantification to estimate film parameters from neutron reflectivity curves.  

\begin{figure}
  \centering
  \includegraphics[width=0.88\textwidth]{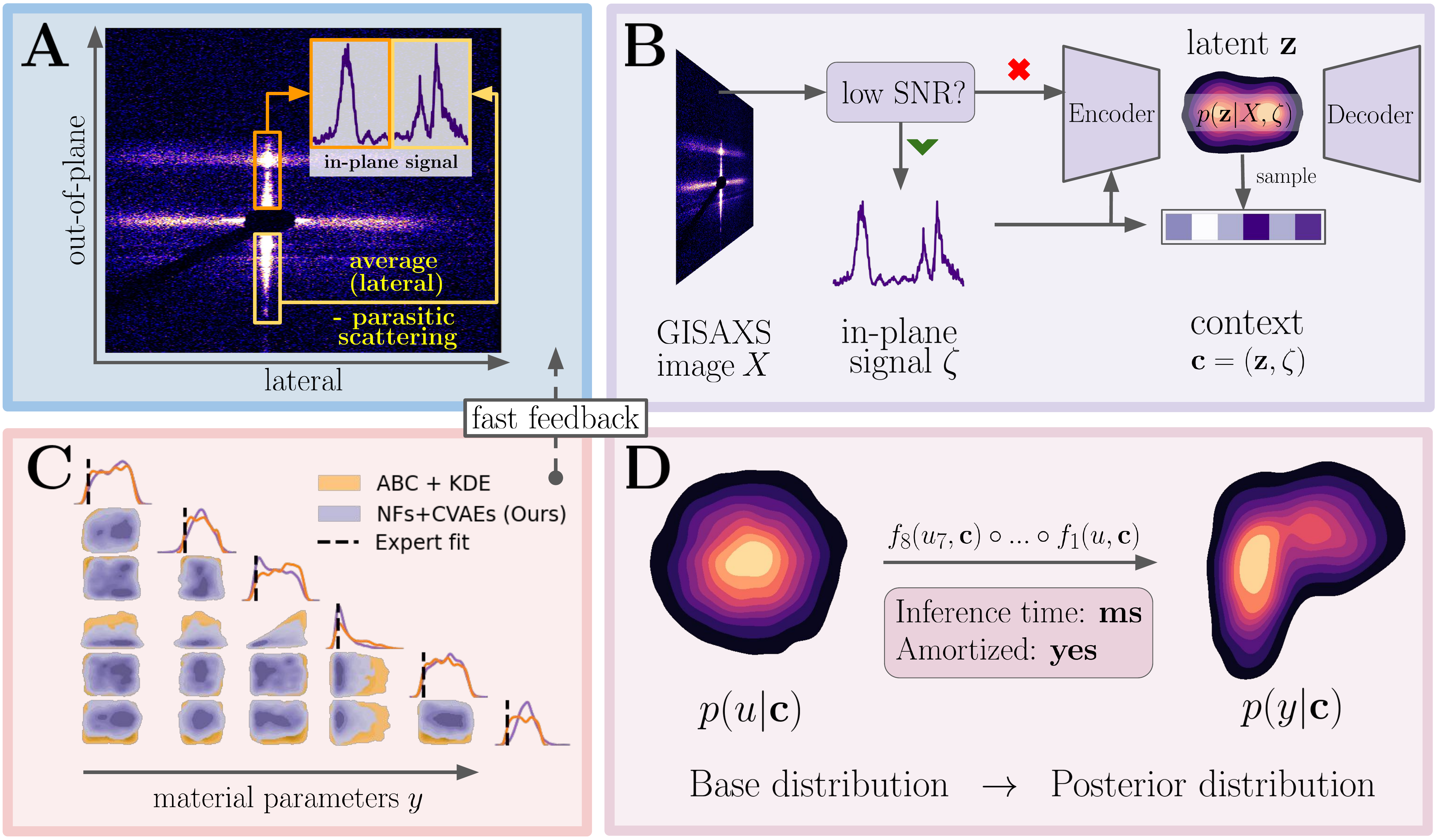}
  \caption{A schematic overview of the proposed inference pipeline for GISAXS data (A). In the first step, conditional probability $p(X, z|\zeta)$ over GISAXS data $X$ and latent variables $z$ given the in-plane GISAXS signal $\zeta$ is approximated by conditional VAEs (B). The probabilistic model allows us to compute robust representation $\mathbf{c}$ that can be obtained even when only $\zeta$ is given as input. Afterwards, we approximate posterior distribution $p(y|\mathbf{c})$ over object parameters with normalizing flows (D), yielding fast inference that allows accelerating feedback during experiments (C).}
  \label{fig:pipeline}
  \vspace{-1em}
\end{figure}

\paragraph{Our contribution}
In this paper, we develop an inference framework allowing for fast and robust reconstruction of GISAXS data to accelerate GISAXS data analysis. As the signal-to-noise ratio (SNR) of experimental images might suffer from distortions caused by grazing-incidence geometry \cite{Liu2018}, we mainly focus on using an in-plane scattering signal\footnote{We define the \emph{in-plane scattering signal (profile)} as the average of a central region (see Fig. \ref{fig:pipeline}A) of an image over the lateral dimension of a detector with subtracted parasitic scattering signal from the beamstop.} as input. Despite the recent success of discriminative neural networks, such models do not account for the inherent ambiguity of reconstruction and, as a rule, do not provide a researcher with uncertainty quantification. Instead of learning a function from images to parameters, we use the Bayesian approach and estimate the posterior distribution of object parameters given the GISAXS data. Our framework has a two-fold structure. In the first step, we learn a robust probabilistic representation of the GISAXS generative process with variational auto-encoders \cite{Kingma2014}. Second, we model the posterior distribution via likelihood-free inference \cite{Cranmer2020} with normalizing flows \cite{Rezende2015}.  


\section{Methods}

\paragraph{Conditional variational auto-encoders}
Variational autoencoders (VAE) \cite{Kingma2014} is a framework to model the data generation process via approximation of joint probability $p(x,z)$ over observed variables $x$ and latent variables $z$. It combines a generative model $p_\theta(x|z)$, an inference model $q_\phi(z|x)$ and a prior $p(z)$, allowing unconditional data generation from a learned distribution model. Conditional VAEs (CVAEs) \cite{Yan2016} is an extension of the framework that models a conditional distribution $p(x,z|y)$. To learn the model parameters $\theta$, $\phi$, one maximizes the conditional log-likelihood $log \: p_\theta(x|y)$ via maximizing the evidence lower bound \cite{Kingma2014, Yan2016}. Subsequently, one can sample from conditional distribution $p_\theta(x|y,z)$ where random noise $z$ attributes for the variance in reconstruction of $x$ from $y$. We model each distribution as Normal distribution with learnable mean and variance.

\paragraph{Normalizing flows} 
We are interested in estimating the intractable posterior distribution of unobserved variable $y$ given an observation $x$. One particular class of neural density estimators that scales well to high-dimensional data is normalizing flows (NFs) \cite{Kingma2014} -  the technique that was applied to inverse problems in various domains, including astronomy \cite{Sharma2022}, lattice field theory \cite{Haan2021}, and high energy physics \cite{Gao2020}. In the framework, a latent random variable $u$ distributed according to a simple base distribution $p(u)$ undergoes a sequence of  $K$ invertible parameterized transformations $z_k = f_k(z_{k-1})$, where $z_0 = u$ and $z_K = x$. The framework can be further generalized to model the posterior distribution $p(x|y)$ by conditioning each function $f_k$ as well as the prior distribution $p_\theta(u|y)$ to the context variable $y$. For a model density $p_K(x|y)$ to approximate the true density $p(x|y)$, the negative log-likelihood of the observed data under the model is minimized.

\paragraph{Proposed framework}
We aim to estimate the posterior distribution $p(y|X)$ of object parameters given corresponding GISAXS data $X$. As the likelihood of the data-generating process is intractable, we employ simulation-based inference \cite{Cranmer2020} via NFs. Due to the scarcity of experimental data, we train the model on synthetic GISAXS data generated by state-of-the-art BornAgain \cite{Pospelov2020} software. Although one can generate arbitrarily many simulated patterns, the distribution of these patterns can alter drastically from the distribution of experimental data. The discrepancy might render the model trained on simulated data inapplicable to real-world data, thus requiring more robust-to-noise data representations such as in-plane GISAXS signals $\zeta$ derived from experimental images $X$. To generalize our framework to both modalities, we use CVAEs to learn joint probability model $p_{\theta_2}(X,z,\zeta)$. Combined, the framework's generative models learn data-generating process whose joint probability distribution is factorized as follows:
\begin{equation*}
   p_{\theta_1, \theta_2}(y, \zeta, z, X) = \underbrace{p_{\theta_1}(y|\zeta, z)}_\text{NFs} \cdot \underbrace{p_{\theta_2}(X,z|\zeta)}_\text{CVAEs} \cdot p(\zeta)
\end{equation*}
where we use the chain rule and assume that $y$ and $X$ are conditionally independent given $\zeta, z$. Therefore, when only the in-plane signal $\zeta$ is available, latent variables are sampled from a prior $p(z)$. Otherwise, one uses GISAXS data to sample $z \sim q_{\theta_2}(z|X, \zeta)$. Parameters $\theta_1$ and $\theta_2$ are optimized by minimizing the sum of loss objectives of NFs and CVAEs. More details on the implementation and optimization procedure are provided in Appendix A.1-2.



\section{Experiments}

\begin{figure}[!tbp]
  \centering
  \begin{minipage}[b]{0.59\textwidth}
    \includegraphics[width=\textwidth]{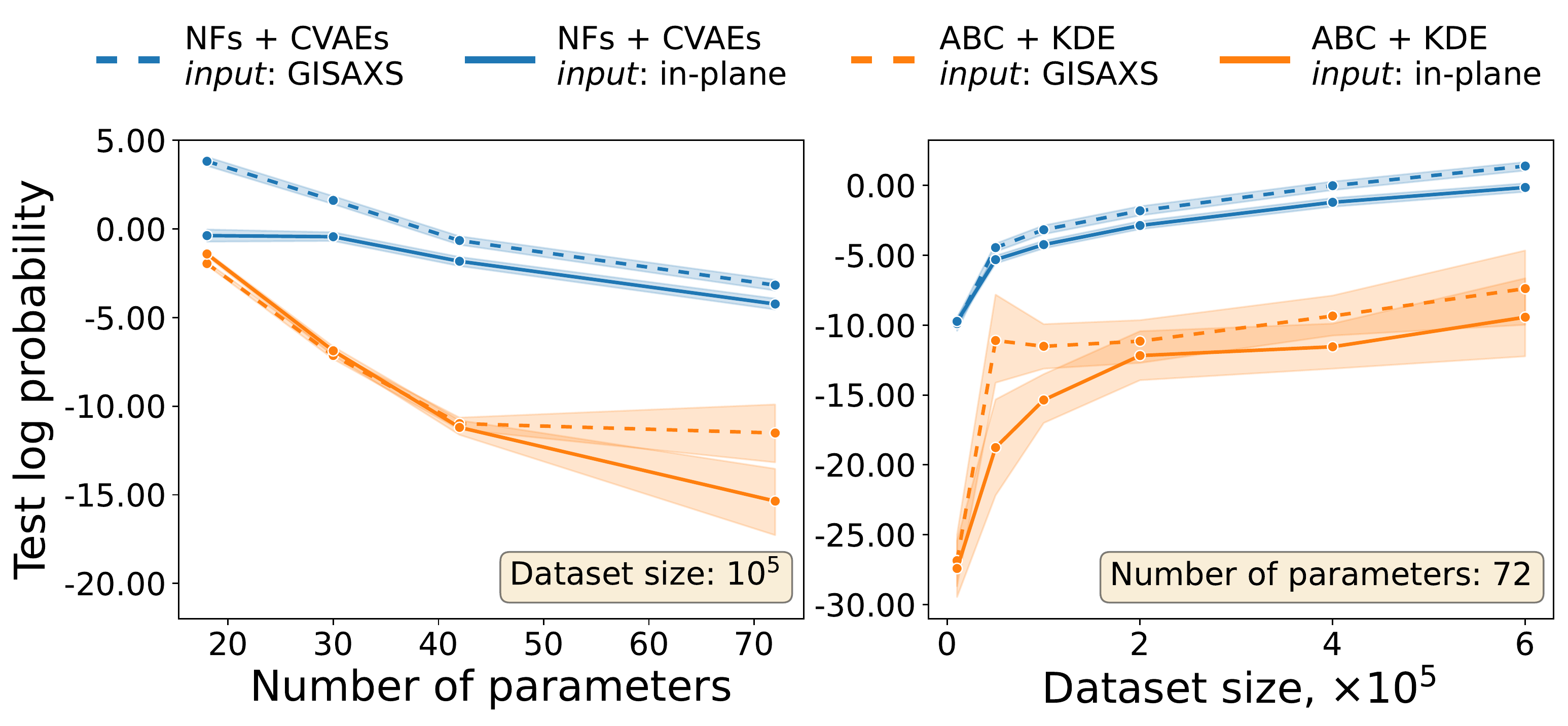}
    \caption{Log probability on test dataset for synthetic GISAXS data. Left: for multiple objects with different complexity. Right: with a varied number of seen data points during training/inference.}
    \label{fig:mat.comp}
  \end{minipage}
  \hfill
  \begin{minipage}[b]{0.36\textwidth}
    \includegraphics[width=\textwidth]{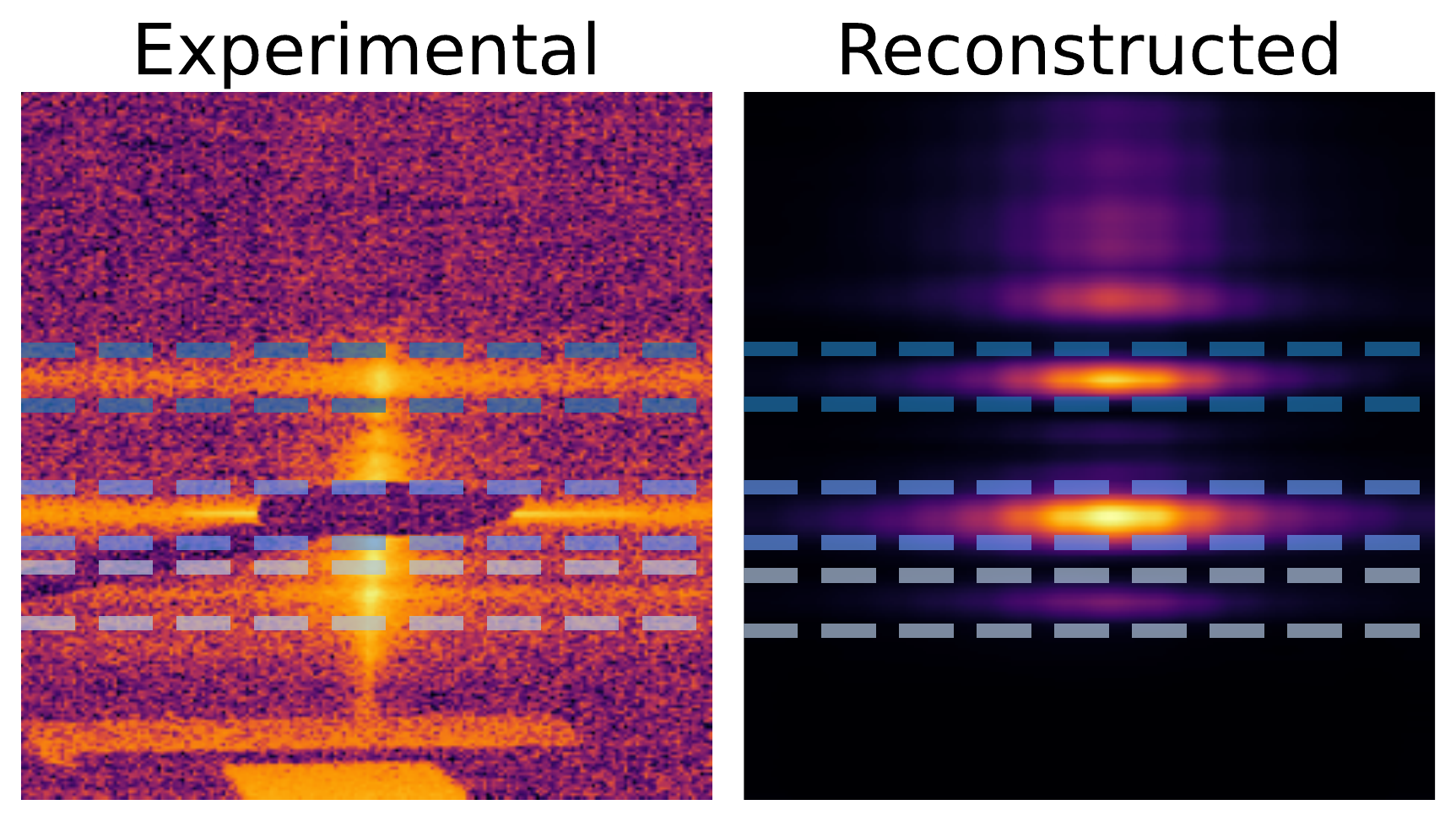}
    \caption{Reconstruction of a GISAXS image from its in-plane signal. Left: ground-truth experimental data. Right: reconstruction by CVAEs. Dashed boxes denote aligning intensity peaks.}
    \label{fig:rec}
  \end{minipage}
  \vspace{-1em}
\end{figure}

\paragraph{Synthetic data}
In this paper, we focus on metallic multi-layer materials consisting of multiple repetitions of Tantalum (Ta) and Copper nitride (Cu$_3$N). The final goal of our research is to apply the inference pipeline to experimental GISAXS data obtained by irradiating a 12-layer sample (TaO - 5$\times$(Ta - Cu$_3$N) - Ta) (see Appendix A.3 for experimental details). We use BornAgain \cite{Pospelov2020} to simulate synthetic labelled GISAXS patterns. We generate $10^5$ images with varied parameters of each layer for 3-layer samples (TaO-Ta-Cu$_3$N), 5-layer samples, 7-layer samples (TaO-Ta-Cu$_3$N-...) and $6 \cdot 10^5$ images for 12-layer samples. Precisely, each layer is described by six parameters: dispersion, absorption, thickness, roughness, Hurst parameter and lateral correlation length. During dataset generation,  we uniformly sample the value of each parameter from a predefined range and simulate a GISAXS image via BornAgain. For each image, we compute the in-plane scattering signal (see Appendix A.4-5 for details on parameter ranges, simulation parameters and preprocessing). 

\paragraph{Baselines and evaluation}
As a baseline approach, we use the combination of kernel density estimation (KDE) (Gaussian kernel, $0.2$ bandwidth) and approximate Bayesian computation (ABC) \cite{Cranmer2020} with Euclidean distance as distance measure and uniform distribution over predefined parameter range as prior. We set the acceptance threshold such that the acceptance rate is equal to $0.002$ on average. To perform inference with ABC, we use precomputed training datasets. As an evaluation metric, we use log (natural) probability and normalized mean absolute error (MAE), i.e. scaled to a dimension range, between object parameters obtained by Randolph et al. \cite{Lisa2022} and parameters with the highest log probability assigned by an inference algorithm.

\paragraph{Material complexity \& dataset size}
We here analyze the influence of material complexity (and hence the number of parameters) on the performance of our inference framework (see Fig. \ref{fig:mat.comp}, left). It can be seen that neural density estimators consistently outperform ABC even with an in-plane scattering signal given as input (NFs+CVAEs). Each algorithm's performance worsens as the number of parameters increases as each layer contributes to the scattering pattern rendering highly complex GISAXS images. Besides, the deeper the layer, the weaker its contribution to the GISAXS image and hence the higher the uncertainty of an estimator. All estimators yield better performance as the number of training data points grows (see Fig. \ref{fig:mat.comp}, right), with NFs + CVAEs demonstrating significantly higher log probability on test data than ABC for any dataset size. 

\paragraph{Experimental results}
\begin{table}[]
\caption{Application of our framework to experimental data. We quantify the deviation of the posterior distribution estimated by our framework from the one estimated by ABC+KDE with Wasserstein distance between marginal 1D distributions averaged across all parameters. We also provide log probability estimated by each framework for parameters obtained by Randolph et al. \cite{Lisa2022} and MAE between these parameters and values with the highest probability attributed by the corresponding framework. We also compute the inference time required for both frameworks for a single CPU\textbackslash GPU.}
\label{tab:results}
\centering
\begin{adjustbox}{max width=0.9\textwidth}
\begin{tabular}{ccccccc}
\hline \\[-0.9em]
           & $W_p(\cdot$, ABC$)  $    &   Log probability        & MAE  & \multicolumn{1}{p{1.5cm}}{\centering Inference \\ time, s} & \multicolumn{1}{p{1.5cm}}{\centering Pretraining \\ time, s} & Amortized \\ \hline \\[-0.9em]
ABC + KDE       & -           & $-22.1^\ast$   & $0.34$ & $18 \cdot 10^6$ $^{\ast \ast}$   & -                   & no         \\ \\[-0.9em]
NFs + CVAEs & $0.06 \pm 0.04$ &  $\mathbf{-14.9}$   & $\mathbf{0.33}$ & $\mathbf{4.56}$            & $18 \cdot 10^6$     & \textbf{yes}                         \\ \hline
\\[-0.9em]
\multicolumn{7}{l}{\small $^\ast$ The average value over multiple bandwidth values ($0.15$-$0.3$) of KDE is calculated.} \\ 
\\[-0.9em]
\multicolumn{7}{l}{\small $^{\ast \ast}$ If the same dataset is re-used each time, the inference time is 45 s.} \\
\end{tabular}
\end{adjustbox}
\vspace{-1em}
\end{table}

We apply our framework to perform inference on experimental data (see Table \ref{tab:results}, Fig. \ref{fig:pipeline}C). The low value of Wasserstein distance between 1D marginal distributions indicates significant overlap between posteriors estimated by NFs+CVAEs and ABC+KDE. Besides, MAE between expert fit and parameters with the highest log probability is slightly smaller for our framework. While the performance of both approaches is similar, the inference of our model is significantly faster. As it is amortized, inference time scales efficiently with the number of observations, while all the steps of ABC have to be repeated for a new observation. Therefore our model can be re-used for multiple observations on large-scale facilities, significantly accelerating data analysis. It is also possible to use CVAEs to reconstruct GISAXS image $X$ from its in-plane scattering signal $\zeta$ (see Fig. \ref{fig:rec}). Here, we sample latent variables $z \sim N(0;\mathbf{1})$ and use decoder to predict image $X$ given $z$ and $\zeta$. One can see that Bragg and Yoneda peaks in both experimental and reconstructed images match well.

\section{Conclusions}
In this work, we utilized a neural density estimator to perform rapid parameter inference on GISAXS data, focusing on data with low SNR when only an in-plane scattering signal is available for inference. We propose a framework based on NFs and CVAEs that estimates object parameters' posterior distribution, given an experimental GISAXS image. We validate our approach on synthetic GISAXS data generated by BornAgain with different pattern complexity and apply it to an experimental GISAXS image obtained from complex multi-layer material. The method demonstrates inference speed-up by over six orders of magnitude compared to the traditional ABC framework enjoying GPU parallelization while yielding a consistent estimated posterior distribution. Furthermore, our method is amortized, meaning it scales efficiently when applied to many observations once trained. We expect the model to be applied to experiments on large-scale facilities that generate many independent and identically distributed observations and ultimately accelerate the analysis of GISAXS data.

\begin{ack}
This work was supported through the project \textit{Artificial Intelligence for Neutron and X-Ray Scattering} (AINX) funded by the Helmholtz AI unit of the German Helmholtz Association. The XFEL experiments were performed at the BL2 of SACLA with the approval of the Japan Synchrotron Radiation Research Institute (JASRI) (Proposal No. 2018B8049).
\end{ack}

\section*{Checklist}

\begin{enumerate}

\item For all authors...
\begin{enumerate}
  \item Do the main claims made in the abstract and introduction accurately reflect the paper's contributions and scope?
    Yes
  \item Did you describe the limitations of your work?
    No
  \item Did you discuss any potential negative societal impacts of your work?
    No
  \item Have you read the ethics review guidelines and ensured that your paper conforms to them?
    Yes
\end{enumerate}

\item If you are including theoretical results...
\begin{enumerate}
  \item Did you state the full set of assumptions of all theoretical results?
    No theoretical results are included.
        \item Did you include complete proofs of all theoretical results?
    No theoretical results are included.
\end{enumerate}

\item If you ran experiments...
\begin{enumerate}
  \item Did you include the code, data, and instructions needed to reproduce the main experimental results (either in the supplemental material or as a URL)?
    Yes, we provide a link to our GitHub repository.
  \item Did you specify all the training details (e.g., data splits, hyperparameters, how they were chosen)?
    Yes, see Appendix A.1-2.
        \item Did you report error bars (e.g., with respect to the random seed after running experiments multiple times)?
    Yes
        \item Did you include the total amount of compute and the type of resources used (e.g., type of GPUs, internal cluster, or cloud provider)?
    Yes, see Appendix A.2.
\end{enumerate}

\item If you are using existing assets (e.g., code, data, models) or curating/releasing new assets...
\begin{enumerate}
  \item If your work uses existing assets, did you cite the creators?
    Yes
  \item Did you mention the license of the assets?
    Yes
  \item Did you include any new assets either in the supplemental material or as a URL?
    No
  \item Did you discuss whether and how consent was obtained from people whose data you're using/curating?
    Yes
  \item Did you discuss whether the data you are using/curating contains personally identifiable information or offensive content?
    Yes
\end{enumerate}

\item If you used crowdsourcing or conducted research with human subjects...
\begin{enumerate}
  \item Did you include the full text of instructions given to participants and screenshots, if applicable?
    Not applicable
  \item Did you describe any potential participant risks, with links to Institutional Review Board (IRB) approvals, if applicable?
    Not applicable
  \item Did you include the estimated hourly wage paid to participants and the total amount spent on participant compensation?
    Not applicable
\end{enumerate}

\end{enumerate}


\appendix

\section{Appendix}

\subsection{Model implementation}
Similarly to the model of Mishra-Sharma et al. \cite{Sharma2022}, we employ masked autoregressive flows \cite{Papamakarios2017} as the transformation flow with 8 transformations made of a 2-layer masked autoregressive MLP with a hidden layer size of 128. The model was implemented with \textit{nflows} library \cite{Durkan2020}. Model parameters are optimized by minimizing the following objective:
\begin{equation}
\label{loss:nf}
    \mathcal{L}_{NFs}(x,y; \theta) = - log \: p_K(x|y) = - log \: p(f^{-1}(x, y)) - \sum_{k=1}^K log \: \left| det \frac{dz_k}{dz_{k-1}}\right|  
\end{equation}

To build CVAEs, we use a convolution neural network-based encoder and decoder that are reversed (transposed) versions of one another. Encoder/decoder has 6 blocks where each block consists of a convolutional operator with kernel size 3, batch normalization and SiLU activation. Additionally, we have a conditional encoder - a 2-layer MLP with 128 hidden units. The loss function of CVAEs is defined as follows:
\begin{equation}
\label{loss:cvae}
    \mathcal{L}_{CVAEs}(x,y; \theta, \phi) = - KL(q_\phi(z|x,y) \: || \: p(z)) + \EX_{q_\phi(z|x,y)}\left[\: log \: p_\theta(x|y,z) \:\right]
\end{equation}
where $KL$ denotes the Kullback–Leibler divergence \cite{Kullback1951}. Thus, we minimize the sum of loss objectives to train the parameters of the framework:
\begin{equation}
\label{loss:nfcvae}
    \mathcal{L}_{NFs+CVAEs}(y, \zeta, z, X; \theta_1, \theta_2) = \mathcal{L}_{NFs}(y,(\zeta,z); \theta_1) +  \mathcal{L}_{CVAEs}(X,\zeta; \theta_2)
\end{equation}

Hyperparameters of our framework, such as the number of hidden units and learning rate, were optimized by grid search using \textit{wandb} \cite{Biewald}.

\subsection{Optimization procedure}
We optimize parameters of each model separately from one another by minimizing the corresponding loss objective (Eq. \ref{loss:cvae}, \ref{loss:nf}). We reserve 20\% of training data for validation and 1000 data points for testing. For each model, AdamW \cite{Loshchilov2019} optimizer is used with initial learning rate $10^{-3}$ and weight decay $10^{-2}$. CVAEs are trained for 30 epochs with early stopping and learning rate decaying by the factor of $0.1$ every 10 epochs. We train normalizing flows for 100 epochs with early stopping and decreasing learning rate by $0.1$ every 30 epochs. For each model, the training was performed in mini-batches of size 32. We use NVIDIA Tesla V100 GPU from the Hemera HPC system of HZDR.

\subsection{Experimental data}
We use the same experimental data as Randolph et al. \cite{Lisa2022}. The experimental multi-layer material was grown on a $\SI{700}{\micro\meter}$ thick silicon wafer carrying a Ta seed layer on a $\SI{100}{\nano\meter}$ thick layer of thermal silicon. The experimental GISAXS images have been measured at the EH6 station at the SACLA XFEL facility in Japan \cite{Yabuuchi2019} by irradiating the multi-layer sample with $\SI{8.81}{\kilo\electronvolt}$ X-ray photons under the grazing incidence angle of $\SI{0.64}{\degree}$ and recording the scattered intensity with a 2D-area detector (MPCCD). A beamstop has blocked the strong specular peak as it is orders of magnitude more intense than the diffusely scattered signal. 

\subsection{BornAgain simulation}
BornAgain software uses the DWBA formalism to synthesize GISAXS patterns for a given object and experimental setup. We use multi-layer objects, describing each layer by characteristic properties (see Table \ref{tab:params}). We access BornAgain via Python script, which can be found in our \href{https://github.com/maxxxzdn/gisaxs-reconstruction}{GitHub repository}. We indicate the experimental parameters used in each simulation in Table \ref{tab:exp_params}. Table \ref{tab:params} presents the sample parameters with corresponding ranges from which we sampled uniformly when generating datasets.

\begin{table}[!htb]
    \begin{minipage}{.4\linewidth}
      \caption{Experimental setup}
      \label{tab:exp_params}
      \centering
      \begin{adjustbox}{max width=0.9\textwidth}
        \begin{tabular}{lc}
        Experimental parameters     &                            \\ \hline
        pixel size                  & 50 $\mu$m                  \\
        pixel y-direction           & 512                        \\
        distance in y-direction     & 25.6 mm                    \\
        pixel z-direction           & 1024                       \\
        distance in z-direction     & 51.2 mm                    \\
        sample detector distance    & 1277 mm                    \\
        y-position of specular reflection & 10.75 mm                   \\
        z-position of specular reflection & 20.65 mm                   \\
        wavelength                  & 0.14073 nm                 \\
        incident angle              & 0.64$^{\circ}$    \\
        azimuthal angle             & 0.0$^{\circ}$     \\
        beam intensity              & 10$^{13}$ \\
        constant background         & 60                        
        \end{tabular}
        \end{adjustbox}
    \end{minipage}%
    \begin{minipage}{.55\linewidth}
      \centering
        \caption{Parameter ranges}
        \label{tab:params}
        \begin{adjustbox}{max width=0.9\textwidth}
        \begin{tabular}{lcc}
        Layer parameters                                 & range (Ta) & range (Cu$_3$N) \\ \hline
        dispersion, $\cdot 10^{-5}$ & 1 - 4      & 0.8 - 3         \\
        absorption, $\cdot 10^{-7}$ & 0.1 - 90   & 0.1 - 90        \\
        thickness, nm                                       & 0.3 - 4.7  & 0.3 - 12.0      \\
        roughness, nm                                        & 0.3 - 5    & 0.3 - 5         \\
        Hurst parameter                                  & 0.1 - 0.9  & 0.1 - 0.9       \\
        lateral correlation length, nm                       & 3 - 30     & 3 - 30         
        \end{tabular}
       \end{adjustbox}
    \end{minipage} 

\end{table}

\subsection{In-plane GISAXS signal computation}
To obtain an in-plane signal, we first take only the center cut of 30 pixels along the lateral dimension of a detector around the specular beam as done for the experimental data \cite{Lisa2022}. We then average over the lateral dimension of the detector and split the resulting curve into two parts: before and after the beam stop. Next, we normalize each part and concatenate them yielding an array of length 359.
We subtract an underlying background signal related to parasitic scattering from the beamstop for experimental data. The parasitic scattering model was taken from \cite{Lisa2022}.
\end{document}